\documentclass[10pt,twocolumn,letterpaper]{article}

\usepackage[pagenumbers]{cvpr} % To force page numbers, e.g. for an arXiv version

\usepackage{times}
\usepackage{graphicx}
\usepackage{amsmath}
\usepackage{amssymb}
\usepackage{booktabs}
\usepackage{xcolor}         % colors
\usepackage{graphicx}
\usepackage{array}
\usepackage{xspace}
\usepackage{colortbl} %use in the preamble
\usepackage{newtxtext} % rowan note: txfonts breaks everything :( this is apparently the replacement
\usepackage{relsize}
\usepackage{times}
\usepackage{enumitem}
\usepackage{epsfig}
\usepackage{tabularx}

\usepackage{microtype}
\usepackage[utf8]{inputenc} % allow utf-8 input
\usepackage[T1]{fontenc}    % use 8-bit T1 fonts
\usepackage{newtxtext}

\renewcommand{\spadesuit}[0]{\text{\smash{\raisebox{-1pt}{\includegraphics[height=8pt]{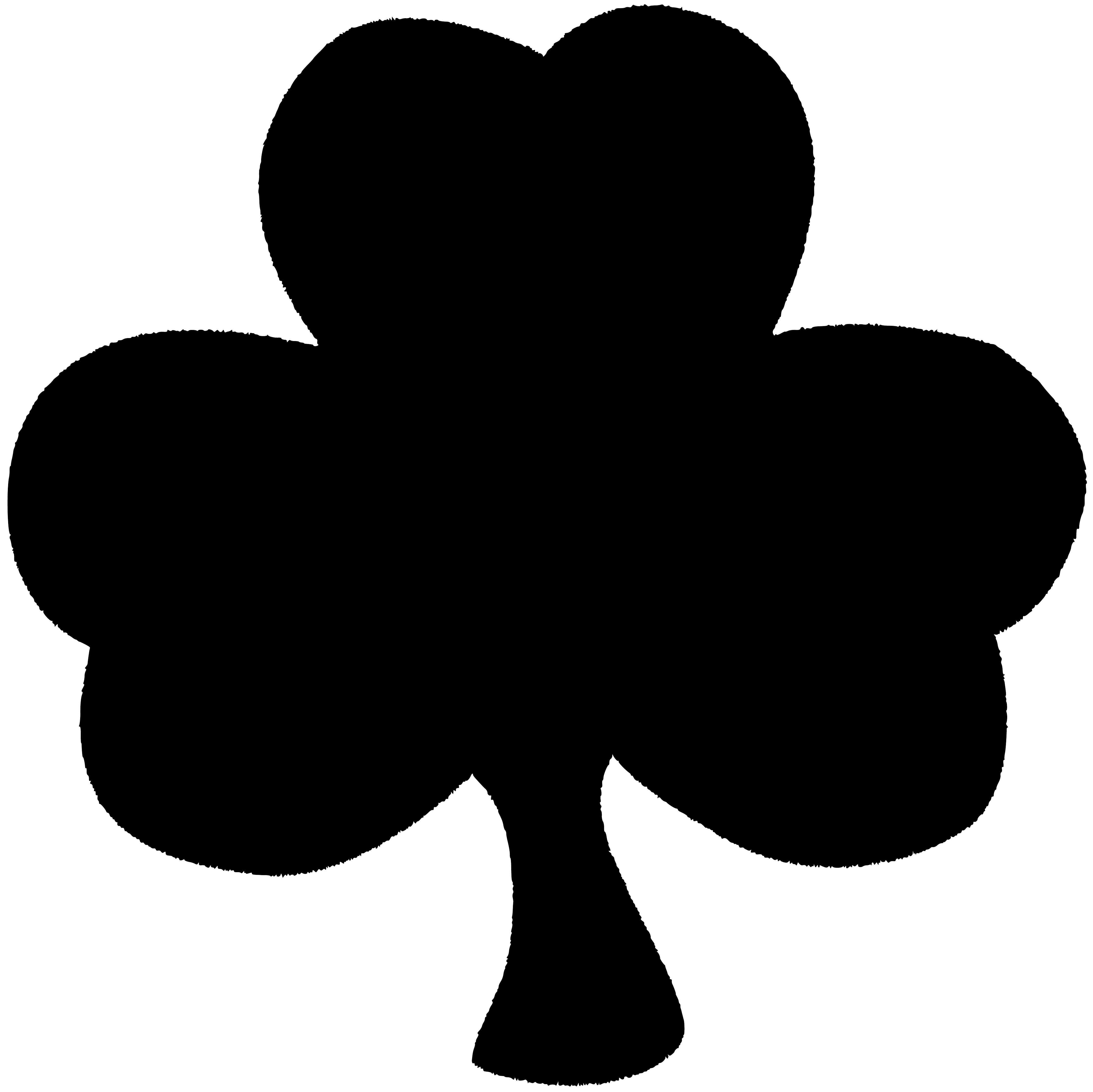}}}}}

\newcommand*{\psyphyfont}{\fontfamily{phv}\selectfont}
\newcommand{\modelName}{{\smash{\small \textbf { {\psyphyfont PERCEP-TL} }}}\xspace}

\usepackage[pagebackref,breaklinks,colorlinks]{hyperref}
\usepackage[capitalize]{cleveref}
\crefname{section}{Sec.}{Secs.}
\Crefname{section}{Section}{Sections}
\Crefname{table}{Table}{Tables}
\crefname{table}{Tab.}{Tabs.}

\definecolor{imagenet-color}{HTML}{2596be}
\definecolor{omniglot-color}{HTML}{444E57}
\definecolor{iam-color}{HTML}{A51E37}

\def\cvprPaperID{***} % *** Enter the CVPR Paper ID here
\def\confName{CVPR}
\def\confYear{2023}

\begin{document}

%%%%%%%%% TITLE
\title{Using Human Perception to Regularize Transfer Learning}

\author{
Justin Dulay and Walter J. Scheirer \\
Dept. of Computer Science and Engineering\\
$^\spadesuit$University of Notre Dame\\
{\tt\small \{jdulay, wscheire\}@nd.edu} \\
}

\maketitle

%%%%%%%%% ABSTRACT
\begin{abstract}
    Recent trends in the machine learning community show that models with fidelity toward human perceptual measurements perform strongly on vision tasks. 
    Likewise, human behavioral measurements have been used to regularize model performance. But can we transfer latent knowledge gained from this across different learning objectives? In this work, we introduce \modelName (Perceptual Transfer Learning), a methodology for improving transfer learning with the regularization power of psychophysical labels in models. We demonstrate which models are affected the most by perceptual transfer learning and find that models with high behavioral fidelity --- including vision transformers --- improve the most from this regularization by as much as \textbf{1.9\%} Top@1 accuracy points. These findings suggest that biologically inspired learning agents can benefit from human behavioral measurements as regularizers and psychophysical learned representations can be transferred to independent evaluation tasks. 
\end{abstract}

%%%%%%%%% BODY TEXT
\section{Introduction}
\label{sec:introduction}
All visual systems process data into latent representations before making decisions. Biological systems receive input through the eyes and encode visual inputs through the optic nerve where it can be processed within the brain through synapses and neural activity. Similarly, artificial vision systems encode image inputs into a functional mathematical space. Images are complex, labels are fuzzy, and features are noisy for many reasons. Both biological and artificial systems capture comparable internalized representations, but a high-fidelity shared representation remains elusive. 

% psychophysics 
One way to capture biological latent representations of data is through psychophysics: the study of systematically changing a stimulus and measuring human response to it.
In the past, studies have modified learning pipelines by incorporating some human behavioral measurements into the loss function~\cite{grieggs2021measuring}, or within the model architecture itself~\cite{huang2022measuring}. Often, human response times towards stimuli yield salient information for a learning model, acting as a form of regularization. Furthermore, psychophysical measurements have been deployed for model architecture search and model evaluation in studying how a model reacts to psychophysical experiments in a similar way to a human~\cite{blanchard2019neurobiological, richardwebster2018visual, zhou2019hype}. This is also akin to alignment, where models that have human-like fidelity are rewarded more.
%otoole, nakayama
A rich field of psychophysical literature exists~\cite{gescheider2013psychophysics, fechner1948elements, o2002recognizing, o1998perception, maljkovic1994priming, nakayama1986serial}, including some work bridging psychophysical experiments and machine learning, 
% but there remains no formal attempt to synthesize various psychophysical experiments as generalized regularization across tasks in computer vision.
but there remains a gap in using psychophysical labels beyond the scope of learning on a singular domain and task.

\begin{figure}[t]
\centering
\includegraphics[width=1\linewidth]{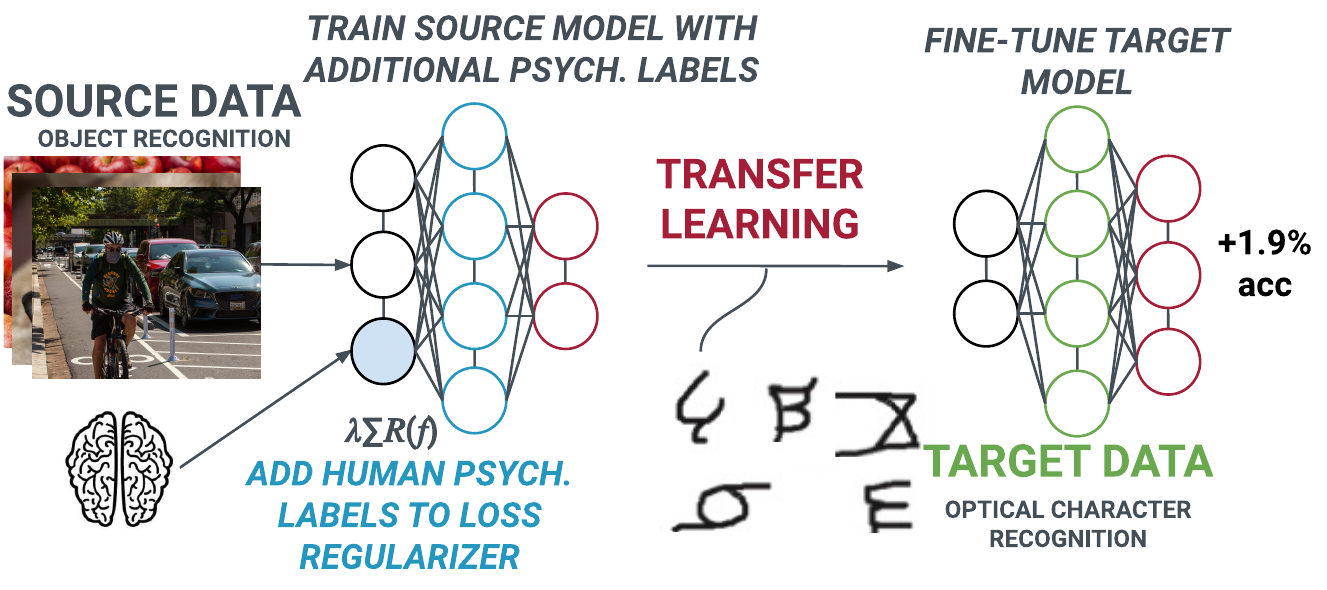}
\caption{\modelName is a model agnostic transfer learning method that uses human behavioral measurements as regularization. Transfer learning, which improves with traditional regularization, often improves furthermore with psychophysical regularization. Overall, the method incorporates human behavioral measurements, such as response times, as additional labels into the regularization term of a loss function during source model training. After this, the model remains more robust to changes in target task distributions (in this case, optical character recognition, but other tasks also transfer). The target model reflects this with increased accuracy on target data and tasks.}
\label{fig:one}
\end{figure}

% transfer learning
To efficiently share learned representations across domains and tasks, transfer learning incorporates the weights of one learned model towards a different learning objective. In recent years, applications of it have seen sweeping improvements in generalist computer vision, as well as in specialties such as medicine and robotics~\cite{deepak2019brain, huynh2016digital, zhu2020transfer}.
% this paper has some theory to use in main part
There have been recent works using regularization within transfer learning~\cite{takada2020transfer}. Moreover, regularization improves most transfer learning tasks.

%concept marriage
% \textcolor{omniglot-color}{
However, there is no work bridging these two core concepts: humans generalize high-fidelity representations of data that can regularize machine learning models, and transfer learning systems improve with regularization. Can we use psychophysically-annotated data to instill strong representations into models such that they can be transferred to other tasks, even \emph{after} fine-tuning? More succinctly, can a model fine-tuned with psychophysically-labeled regularizers improve transfer learning tasks even more than a model trained with only traditional regularizers?

% main point of paper, should i bold or italicize this?
In this paper, we posit that psychophysical labels can help models generalize across tasks: \emph{that when a dataset contains reliable human behavioral measurements, regularizing the loss function towards a more human-centric belief space yields improved model generalization across tasks.} 
% neuro inspired hypothesis
Likewise, we also hypothesize that models on a spectrum of being more biologically inspired tend to also benefit proportionally more from psychophysical regularization than models that are qualitatively less biologically inspired. DiCarlo et. al~\cite{dicarlo2012does} suggest that humans solve latent representations shared among different samples in visual recognition. A model that more closely matches the latent feature representation space of the human brain is indeed ideal, as suggested by Figure~\ref{fig:two}. 

\begin{figure}[t]
\centering
\includegraphics[width=.8\linewidth]{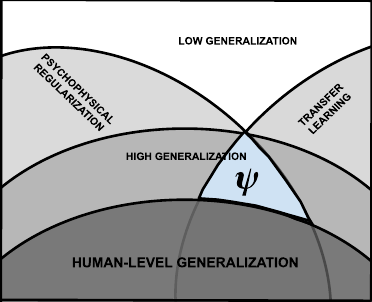}
\caption{\emph{What does it mean to achieve generalization?} Human-level generalization is more profound than any current artificial systems on perceptual tasks. However, many artificial systems can achieve high levels of generalization through various regularizing methods. Artificial systems can achieve better generalization when using human behavioral measurements as additional labels. Likewise, transfer learning generalizes across tasks by transferring latent knowledge from one domain to another. \textbf{Our work} combines these two artificial approaches to achieve better generalization. }
\label{fig:two}
\end{figure}

% summary
In our work, we empirically test these ideas. 
First, we find that models that utilize psychophysical labels within regularizers than their counterparts that do not have these special annotations.
Secondly, we show that transfer learning also improves with models that use psychophysical regularization.

Likewise, we also find that models that are directly \textbf{activation fidelity}, as per qualitative definitions in previous works~\cite{lotter2016deep, lotter2018neural}, often improve performatively when used in conjunction with psychophysical regularizers (though their absolute performance remains is still not as good as traditional deep learning methods). Models with \textbf{behavioral fidelity}, such as vision transformers~\cite{dosovitskiy2020image} with human-like biases~\cite{geirhos2018imagenet}, also perform better with psychophysical regularizers.

Lastly, we compare the models on Brain-Score~\cite{schrimpf2020brain}, demonstrating that several models improving \emph{via} psychophysical labels also show stronger resemblances to state-of-the-art neural activity measurements, albeit not all models.
We find that behavioral fidelity does not always correlate with activation fidelity. 
% In future works, consideration between activation fidelity and actual behavioral fidelity should be accounted for when designing biologically-inspired neural models.

% contributions
Our main contributions are: 
\begin{enumerate}[label=(\textbf{\alph*)}]
    \itemsep0em
    \item Psychophysical labels, when added to the regularization terms in loss functions, improve model generalization.
    \item Transfer learning among different vision tasks improves when using models pre-trained with psychophysical labels and regularization. 
    \item We show Brain-Score~\cite{schrimpf2020brain} evaluations of regularized models with neural activity.
\end{enumerate}

Our results show promise in transfer learning and regularization, which are fundamental concepts in machine learning research. Psychophysical studies unlock potential for more resilient learning representations in future machine learning research.~\footnote{Our code will be released after publication.}

\section{Related Work}
\label{sec:related_works}
\textbf{Transfer Learning.}
Transfer learning is the knowledge acquisition of a source domain and task and applying it through a learned model on a target domain and task~\cite{torrey2010transfer}.  Over the past several decades, transfer learning has gained immense traction, in particular, after deep learning became fashionable~\cite{zhuang2020comprehensive}. With supervised learning, niche domains include medicine (\emph{e.g.} transferring domain knowledge of radiological samples from many adult patients to few pediatric patients~\cite{chouhan2020novel}) and simulation-to-reality transfer in robotic applications~\cite{zhao2020sim}. In the mainstream, many researchers use transfer learning when using an ImageNet pre-trained PyTorch model~\cite{NEURIPS2019_9015} or when efficiently fine-tuning BERT on different corpora~\cite{houlsby2019parameter} (among many other examples).
% then more general cases, original workshops, imagenet, NLP, biological fidelity, regularization

Similar to psychophysics, transfer learning began from experimental psychology. 
In 1901, Thorndike and Woodworth introduced the \emph{transfer of practice}, that transfer from one domain to another was as good as the similarities between the domains~\cite{orata1928theory}. Likewise, acclaimed psychologist B.F. Skinner considers transfer learning a form of generalization~\cite{skinner2014contingencies}. We take a similar approach in our work by complementing generalization in transfer learning with regularization from psychophysics. 

\textbf{Psychophysics.} 
% psychophysics works (all of the stuff in our previous paper)
Psychophysical measurements of human behavior represent a richer source of information for supervised machine learning. 
Many psychophysical experiments collect human perceptual responses through carefully designed experiments in response to varying stimuli. While psychophysics originated long ago~\cite{fechner1948elements}, research in machine learning communities have demonstrated promising results incorporating ideas inspired by it into machine learning paradigms. 

Scheirer et al.~\cite{scheirer2014perceptual} introduced a method for incorporating reaction times of psychophysical measurements into a support vector machine loss function. In their labeling task, their crowd-sourced annotator selected items \emph{via} an alternative forced choice task, where the response time was recorded for each action and subsequently added as an additional label for each class. 
These techniques have been applied in a variety of domains, such as affective recognition \cite{mccurrie2017predicting, ponce2016chalearn}, robotics \cite{milford2019self, zhang2018agil}, reinforcement learning~\cite{guo2021machine}, and human document transcription \cite{grieggs2021measuring}, among others~\cite{webster2018psyphy, dicarlo2012does, boyd2021cyborg, seer2022}. Likewise, there is a growing focus in the literature to bridge gaps of robotic learning systems towards generalized intelligent agents\cite{grauman2022ego4d, ramakrishnan2021environment, min2021integrating, fathi2012learning} --- all point towards generalization as a concept. In our work, we use psychophysical labels as a regularizing tool; that is, if psychophysical regularizers help learning systems generalize, then they also should help transfer learning tasks. 

%ViT
\textbf{Scaling Neural Architectures Upon Biological Fidelity.} 
Since their inception, neural models have had various degrees of reference to biological fidelity. Convolutional neural networks are loosely inspired by biological networks, yet score high on heuristics associated with \textbf{biological fidelity}~\cite{schrimpf2020brain}. However, vision transformers, while not explicitly stated to be inspired by the human brain, have more similar biases to humans than convolutional neural networks do~\cite{geirhos2018imagenet}. Likewise, other networks, such as predictive coding networks, are directly biologically-inspired with high \textbf{activation fidelity} but maintain poor performance on learning tasks. In our work, we explore how these models, on varying degrees of behavioral fidelity, perform when using them with psychophysical regularizers.

Vision transformers~\cite{dosovitskiy2020image} have seen sweeping performance enhancements on benchmarks in recent years, along with some interesting comparisons to human performance on human tasks~\cite{tuli2021convolutional}. They have shown promising results on a variety of tasks including image captioning~\cite{liu2021cptr, radford2021learning} and feature representation~\cite{liu2021swin}, tasks that have been utilized in previous psychophysics and machine learning studies, among others~\cite{han2022survey}. 

Vision transformers appear to have different implicit biases than convolutional neural networks. Tuli et al.~\cite{tuli2021convolutional} suggested that they may be biased more towards shapes, as opposed to textures~\cite{geirhos2018imagenet}, which is a human trait~\cite{tuli2021convolutional} and implores higher behavioral fidelity than other methods. Other research has also suggested that Vision Transformers' learned representations are different than convolutional neural networks and remain open to further study~\cite{raghu2021vision, naseer2021intriguing, steiner2021train}.

Predictive coding is a theory of learned representations in humans~\cite{rao1999predictive}, where the brain updates a working model of its perceptual representation over time.
Directly inspired by neurological predictive coding, PredNet was introduced as an artificial imitation through a series of convLSTM recurrences over time steps, encoding temporal information~\cite{lotter2016deep}. This model is high in activation fidelity with a direct attempt to model a psychological theory on predictive coding of neurological signals~\cite{rao1999predictive}. While the original task was frame prediction, object recognition~\cite{wen2018deep}, pose estimation~\cite{banzi2020learning}, and object-matching have been researched~\cite{blanchard2019neurobiological} with it --- all of which are suitable for transfer learning between tasks.

\textbf{Our work} marries the combination of transfer learning and psychophysics as complementary generalization components.

\section{Core Concept: \modelName}
\label{sec:psychophysical_transfer_learning}
In this section, we present \modelName, including: 

\begin{enumerate}[label=(\textbf{\alph*)}]
    \itemsep0em
    \item How transfer learning occurs among tasks in this paper~\ref{sub:transfer_learning}.
    \item Psychophysical regularization for fine-tuning models~\ref{sub:regularization_in_psychophysics}.
    \item The datasets that we used in our experiments~\ref{sub:datasets}.
    \item Justification of psychophysical regularization in the context of transfer learning~\ref{sub:why_use_this}.
\end{enumerate}

\subsection{Transfer Learning}
\label{sub:transfer_learning}

\modelName is the utilization of psychophysical labels in the regularization term of the loss function during model fine-tuning and transfer learning with this regularized model.
This process is flexible in that it remains agnostic to initial training sets and models, meaning that, for example, a model can be trained on an object recognition task, fine-tuned with psychophysical labels, then transfer learn on a handwriting task. It trains with traditional supervised learning models can also be evaluated on unsupervised learning tasks.

More formally, \modelName consists of a training stage component and a transfer learning component:

\begin{enumerate}[label=(\textbf{\roman*)}]
\item Pre-train model $\theta_p$ on prior source domain $\mathcal{D}_p$ on task $\mathcal{T}_s$ \emph{(optional)}.
\item Train a model $\theta_s$ on source domain $\mathcal{D}_s$ on $\mathcal{T}_s$.
\item Fine-tune $\theta_s$ with additional psychophysical labels $\mathcal{Y}^\psi$.
\item Transfer learned model $\theta_s^\psi$ $\rightarrow$ $\theta_t^\psi$ on target domain $\mathcal{D}_t$ on task on $\mathcal{T}_t$. 
\end{enumerate}

\begin{figure*}[t]
\centering
\includegraphics[width=1\textwidth]{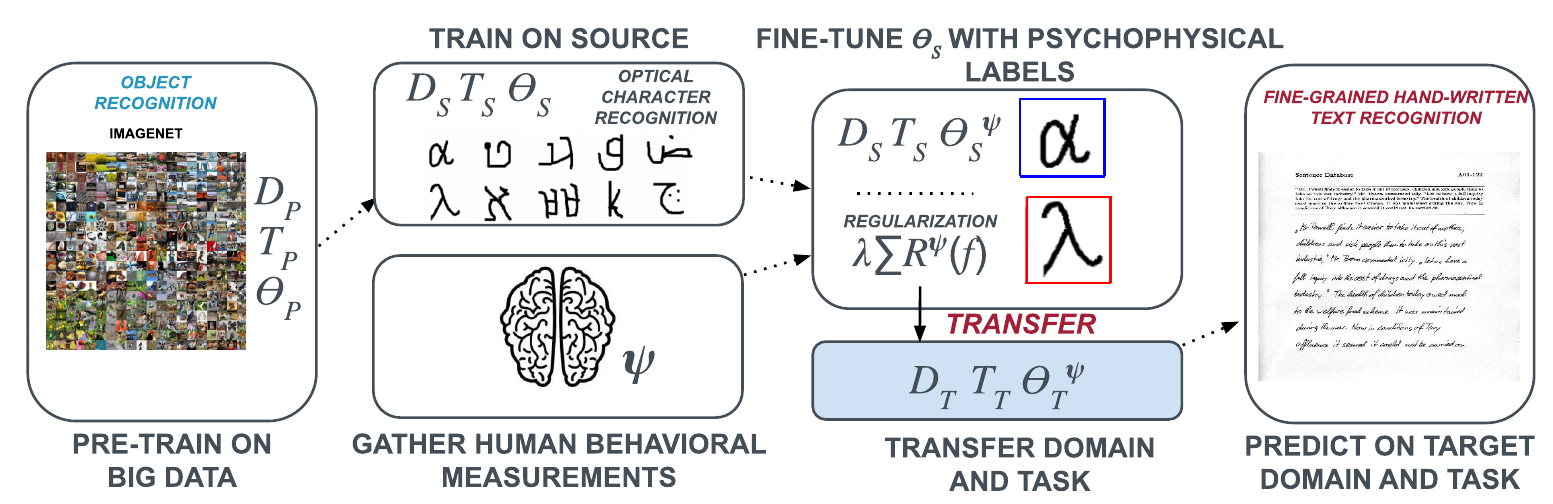}
\caption{
We visualize a neural model $\theta$, a domain $\mathcal{D}$, and a learning task $\mathcal{T}$. In the \emph{optional} left panel, the three components operate in the pre-training stage $p$. In the second column of panels, $\theta_s$ is trained on the source domain $\mathcal{D}_s$ with source task $\mathcal{T}_s$, while \textbf{human behavior measurements $\psi$} are also gathered in the same source domain $\mathcal{D}_s$ through \textbf{regularizing} the loss function with regularization term $R$. Next, $\theta_s^\psi$ is fine-tuned in the source domain with additional $\psi$. This latent knowledge is transferred, as in typical transfer learning, to target domain, task, and model $t$.
}
\label{fig:three}
\end{figure*}

The superscript $\psi$ represents psychophysical labels (\emph{e.g.} reaction times) gathered from human behavioral measurements which are used in fine-tuning of the source model $\theta_s$. With these general steps, we create a framework for transferring domain knowledge from one task to another with most models. Note, not all models need be pre-trained, but many are originally pre-trained on ImageNet. Every domain $\mathcal{D}$ represents a dataset, and every task $\mathcal{T}$ represents a learning task (\emph{e.g.} multi-class classification). These steps are also outlined in Figure~\ref{fig:three}.

% attempt at an example, may remove the short prelims if this fits in okay 
For example, in an object recognition task, psychophysical labels $\psi$ are gathered from human response times to correct identify an object from some noise.  $\psi$ is included in the regularization term of the loss function during the fine-tuning of a model $\theta_s$ on the same source object recognition data. From there, the psychophysically learned model can be used on a target task like traditional transfer learning, but with additional psychophysical regularization as \modelName.

\subsection{Regularization with Psychophysical Labels}
\label{sub:regularization_in_psychophysics}

\textbf{Preliminaries.}
Regularization prevents a model from being biased towards out-of-distribution samples. In explicit regularization cases, the loss function of the model is influenced by a regularization term. In general, this serves as a decorator for any simple loss function, and thus remains model agnostic. 

A generic loss regularization appears as: 

\begin{equation}
\mathcal{L} = \frac{1}{n} \left( \sum_{i=1}^{n} L \left(y_{i}, f\left(x_{i}\right)\right)+ \lambda \sum_{i=1}^{n} R(y_i, x_i) \right)
\label{equation:one}
\end{equation}

In the above Equation~\ref{equation:one}, the loss output space $\mathcal{L}$ results from the base prediction function, given some input $x_i$ and a ground truth label $y_i$ at some batch step $i$ in feature space $x\in \mathcal{X}$ and label space $y \in \mathcal{Y}$. 
For simplicity, $L$ is a generic loss function. 
Likewise, $R$ is a generic regularization function, remaining model agnostic as a function of the inputs and outputs. 
% The resulting logit space of the loss function $L$ follows a logit space $\mathcal{L}=P(y,f(x))$. 
Regardless of the definition of $L$ or $R$, $\lambda$ is a regularization constant that constrains the loss space $\mathcal{L}$. 

% this section can be cut down some if necessary, but not sure how to do it
\textbf{Common Regularizers.} The most common forms of model regularization are $\ell_{1}$ and $\ell_{2}$-regularizers. Both are used extensively within the machine learning community for their ability to create a smoother output space for models and increased generalization capacity~\cite{han2022survey, sun2019survey}. 

Likewise, it remains contested which to use at a given time~\cite{wei2020implicit}. Because of this contest, we believe that we can improve model regularization with a new regularizer that also combines human annotations.

$\ell_{1}$-regularization takes the mean absolute error of the \emph{direct} outputs of the loss function $f$ and applies the absolute difference in between terms: $\ell_{1} = \sum_{i}^{n}|x_i|$. Likewise, the $\ell_{2}$-regularization remains similar to the aforementioned, but \emph{squares} the logits: $\ell_{2} = \sqrt{\sum_{i}^{n}(x_i)^2}$. 

% Dropout
Dropout~\cite{srivastava2014dropout} removes weighted outputs for some layer in $\mathcal{L}=P(y,f(x))$. If the weights fall within a Bernoulli distribution such that they may be pruned, their answer is $1$ in $k \sim Bernoulli(p)$, where $p$ is a scalar hyperparameter for skewing the distribution. 

$\ell_{1}$-regularization, $\ell_{2}$-regularization, and Dropout reduce model complexity and overfitting.

% \modelName as a reg ...
\textbf{Formulation.}
Here, we discuss the formalization of \modelName in training regimes. 
Specifically, we use multi-class cross entropy loss with a modified generalized regularization term:

% Cross entropy 
\begin{equation}
    \mathcal{L} = -\left(\sum_j y_j \log(\hat{y}_j) + \lambda \sum_j R(w_j) \cdot \psi_j )\right)
\label{equation:two}
\end{equation}

The left term is cross-entropy loss with $\hat{y}$ as the predicted logits at $j$. $R$ is a general regularization function on the logits $w_j \in \mathcal{L}$. $\psi_j$ is a psychophysical regularization constant at data point $j$. 
Also, $\psi_j$ is psychophysical measurement of human behavior. For example, $\psi_j$ of a reaction time is the difference between the maximum reaction time of a human at a sample by the actual recorded reaction time at the sample.
Following this, $R$ is multiplied by $\psi$ in cases in which the data receives a psychophysical penalty:
% the easy-vs-hard part, the right-vs-wrong, given our previous works

\begin{equation}
\psi_j =
    \begin{cases}
        w_j \cdot c & \text{if } \hat{y}_j \neq y_j\\
        w_j & \text{otherwise}
    \end{cases}
\label{equation:three}
\end{equation}

with $c$ is a constant scaled penalty within the data. $R$ is specific in application; for example, it can be $\ell_1$.

\subsection{Datasets}
\label{sub:datasets}

We used three psychophysical datasets for this paper. The first dataset uses a variant of ImageNet~\cite{deng2009imagenet}; the second uses a variant of Omniglot~\cite{lake2015human}; the third is a variant on the IAM handwriting dataset~\cite{marti2002iam}. More detailed descriptions can be found in the supplementary material. 

\textcolor{imagenet-color}{\textbf{Psych-ImageNet}}
is a human-annotated dataset from a modification of ImageNet~\cite{deng2009imagenet} with 293 total classes. Each data point has a psychophysical label (reaction time), class label, and ImageNet-sized (224x224) image associated with it.

% description of omniglot set
\textcolor{omniglot-color}{\textbf{Psych-Omniglot}}
consists of a subset of the Omniglot dataset~\cite{lake2015human} with psychophysical annotations. The dataset contains images of handwritten characters from hundreds of typesets, many of which a typical crowd-sourced study participant would be unfamiliar with. The data is augmented with counterpart samples for each image with a deep convolutional generative adversarial network (DCGAN)~\cite{goodfellow2014generative} to increase intraclass variance and the sample size per class --- all of which are forms of implicit regularization.

\textcolor{iam-color}{\textbf{Psych-IAM}}
is a psychophysically-annotated version of the original IAM dataset~\cite{marti2002iam}. This dataset is similar to the annotations collected by Grieggs et. al in ~\cite{grieggs2021measuring}, but with a smaller subset of the data. In our work, we used 2,000 text lines with reaction time annotations. 

Each annotation was collected for a character recognition task and word recognition task. A timer recorded the reaction time (response time) for the annotator to perform each of these tasks respectively. We report results for these three categories in the Experiments Section~\ref{sec:experiments}.

\subsection{Why Use \modelName ?}
\label{sub:why_use_this}

\textbf{Model Agnosticism.} 
\modelName interfaces with a variety of machine learning models. In our work, we modify cross entropy loss variants and sophisticated predictive coding loss formulations, but the regularizer is compatible with other loss functions and models. It can substitute $\ell_{1}$-regularization in many situations. In our work, we experiment with the efficacy of this on deep convolutional neural networks, vision transformers, and deep predictive coding networks, but the regularizer can interface with any loss function that can also with a regularizer, which is in most supervised learning situations~\cite{han2022survey, bassily2018model, finn2017model}.  

\textbf{Generalizability.}
Regularizers, in general, reduce model overfitting towards spurious samples. In the case of \modelName, the outlier samples are ones where the model guesses incorrectly in a case where some human annotator guessed correctly with short reaction time. The penalty produced by the \modelName logits \emph{increases} when the model gets \textcolor{blue}{easy} examples wrong and \emph{decreases} when the model guess correctly on \textcolor{red}{hard} examples.

Furthermore, while transfer learning improves with regularization~\cite{takada2020transfer}, it improves more with psychophysical regularization. The results in this paper demonstrate increased generalization capacity with it.

\textbf{Availability.}
Psychophysical annotations scale with models and datasets. In our experiments, we pre-train PredNet on KITTI~\cite{geiger2013vision}, and evaluate this using psychophysical annotations from the \textcolor{imagenet-color}{Psych-ImageNet} dataset. In the other experiments, we incorporate the annotations directly into the model training regime. 
This shows flexibility to \textbf{(1)} train the model directly and \textbf{(2)} evaluate different models \emph{and} datasets.
Data collection for crowd-sourced human psychophysical studies is relatively easy, as seen in subsection~\ref{sub:datasets}.

\section{Experiments}
\label{sec:experiments}
In this section, we detail the experiments for \modelName. In \ref{sub:train_model_configurations}, we detail the configurations of each of the models trained with psychophysical regularizers and the transfer learning tasks among them. In ~\ref{sub:regularization_ablations}, we detail our ablations of different regularizers.  In~\ref{sub:transfer_learning_eval}, we analyze these experimental results among transfer learning tasks. Lastly~\ref{sub:brain-score}, we evaluate each transfer learned model with Brain-Score~\cite{schrimpf2020brain}, a statistical framework that tests the biological fidelity of a model to neural activity. 

\subsection{Train Model Configurations}
\label{sub:train_model_configurations}

Below are the training model configurations. For each model, we selected a standard cross-entropy loss as a control measurement. Against this, we ran experiments with $\ell_{1}$-regularization, $\ell_{2}$-regularization, Dropout~\cite{srivastava2014dropout}, $\psi$, and $\psi+$Dropout as experimental trials. Each trial ran the same $5$ seeds, and the error bars were calculated \emph{via} standard error.

\textbf{CNN.}
In this, we configured runs with ResNet-50~\cite{he2016deep} and VGG-16~\cite{simonyan2014very}, using the behavioral models as the Tuli et. al~\cite{tuli2021convolutional}. Both models utilized the PyTorch repository~\cite{NEURIPS2019_9015} and pre-trained on ImageNet-21k~\cite{deng2009imagenet}.  

We use traditional CNN models because of their high usage in benchmark scores over the past decade in research~\cite{goodfellow2016deep}, alongside previous experiments combining psychophysics and ML concepts. We further some of these approaches by showing generalizations to other models, as well. 

\textbf{ViT.}
Vision transformers (ViT)~\cite{dosovitskiy2020image} are known to have biases towards shape, more so than textures like CNN's do~\cite{tuli2021convolutional}. Because they have different inductive biases than CNNs, ViTs may be affected differently by using psychophysical labels in their loss functions that regulate their latent feature representations. In our work, we are interested in seeing whether a ViT is affected more proportionately than normal CNNs when using psychophysical labels. 

In our experiments, we use the pre-trained ViT-L from HuggingFace~\cite{huggingface}. The model is pre-trained on ImageNet-21k. We freeze the weights on the last layer and replace them with a linear classifier with the total number of classes (293) output gates. For each of these, we then apply the control cross entropy loss in conjunction with one of the experimental regularizers. 

\textbf{Psych-IAM Experiments.} 
In our experiments with \textcolor{iam-color}{Psych-IAM}~\ref{tab:two}, we utilized compared trials between a convolutional recurrent neural network \emph{via} \modelName, and a visual transformer decoder ~\cite{vaswani2017attention, dosovitskiy2020image} encoder with the same loss modifications. 

The \textcolor{iam-color}{Psych-IAM} contains metrics on character recognition and overall word recognition.
Because this dataset required different metrics than overall Top@1 accuracies, we reported resultant terms in Character Error Rate (CER) and Word Error Rate (WER), respectively, as described in ~\ref{sub:datasets}.

\textbf{PredNet.}
We use a deep predictive coding network (PredNet~\cite{lotter2016deep}) pre-trained on video frames from the KITTI dataset~\cite{geiger2013vision}, on the cities and roads subset. We used the weights saved from this network for a prediction task on the human-annotated datasets. 
% The task of the evaluative model was to predict a frame given the weights, class input, and \modelName information determining whether the class was \textcolor{red}{\emph{easy}} or \textcolor{blue}{\emph{difficult}}. Again, the easy samples the model classifies incorrectly are weighted more heavily in \modelName than hard examples. 

 PredNet outputs are frame-by-frame activations on the inputs. The activations from the first layer of PredNet were used to make a class prediction on the dataset. Therefore, the PredNet that we used was first pre-trained on KITTI, then evaluated as a transfer learning model on its first layer outputs to compare with the other models used in this experiment. 

\begin{figure}[t]
\centering
\includegraphics[width=1\linewidth]{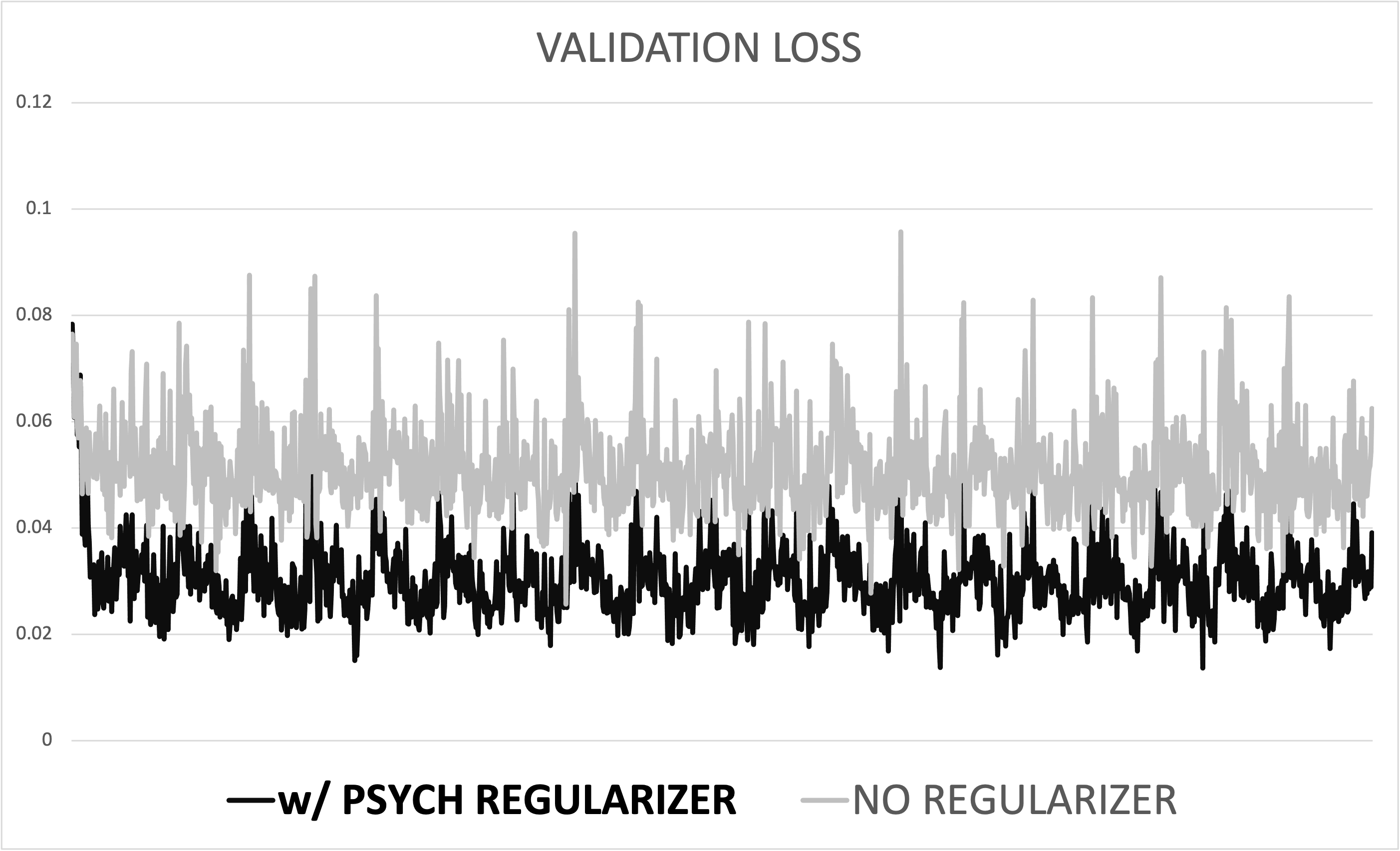}
\caption{\textbf{Validation loss} of a PredNet model on \textcolor{imagenet-color}{Psych-ImageNet}. The top curve does not use the psychophysical labels, while the bottom curve incorporates them into the regularization term $R$ of the loss function. }
\label{fig:four}
\end{figure}

\subsection{Regularization Ablations}
\label{sub:regularization_ablations}
%overview
Here, we present the ablations of regularizers on different neural models. 

\textbf{Observations.}
In our results, we observe that training error decreases more readily when using psychophysical labels in conjunction with regularizers \emph{vs} the standard regularizers in most cases. 
% Qualitatively, we observe better frame predictions when PredNet uses \modelName. 

% if you want to, you can plug in MAE instead of top1 here, since you have labels for columns
\begin{table*}
% {\textwidth}
  \begin{center}
    {\small{
    
    % \begin{tabularx}{lcccc|cccc}
\begin{tabularx}{\textwidth}{XXXXX|XXXX}
&& \textcolor{imagenet-color}{Psych-ImageNet} &&&& \textcolor{omniglot-color}{Psych-Omniglot} \\

 & Top@1 & Top@1 & Top@1 & Top@1 & Top@1 & Top@1 & Top@1 & Top@1  \\

% \textcolor{imagenet-color}{Psych-ImageNet} & Top@1 & Top@1 & Top@1 & \textcolor{omniglot-color}{Psych-Omniglot} & Top@1 & Top@1  \\
\toprule
Method           & ResNet-50       & VGG-16             & ViT    & PredNet         & ResNet-50       & VGG-16             & ViT & PredNet \\
\midrule
Control          & 0.70 $\pm$ 0.05 & 0.72 $\pm$ 0.05 & 0.74 $\pm$ 0.03 & 0.59 $\pm$ 0.03 & 0.78 $\pm$ 0.04 & 0.77 $\pm$ 0.05 & 0.81 $\pm$ 0.04 & 0.63 $\pm$ 0.02 \\
$\ell_{1}$  & 0.71 $\pm$ 0.01 & 0.72 $\pm$ 0.03 & 0.76 $\pm$ 0.03 & 0.62 $\pm$ 0.02 & 0.78 $\pm$ 0.03 & 0.78 $\pm$ 0.05 & 0.81 $\pm$ 0.03 & 0.65 $\pm$ 0.05 \\
$\ell_{2}$  & 0.72 $\pm$ 0.02 & 0.73 $\pm$ 0.02 & 0.76 $\pm$ 0.02 &  0.61 $\pm$ 0.03 & 0.77 $\pm$ 0.04 & 0.77 $\pm$ 0.04 & 0.81 $\pm$ 0.04 & 0.64 $\pm$ 0.03 \\

Dropout  & 0.72 $\pm$ 0.02 & 0.73 $\pm$ 0.02 & 0.76 $\pm$ 0.02 &  0.61 $\pm$ 0.05 & 0.77 $\pm$ 0.04 & 0.77 $\pm$ 0.03 & 0.81 $\pm$ 0.04 & 0.66 $\pm$ 0.05 \\

$\ell_{1}$+Dropout  & 0.73 $\pm$ 0.02 & 0.74 $\pm$ 0.05 & \textbf{0.79} $\pm$ \textbf{0.02} & 0.62 $\pm$ 0.04 & 0.77 $\pm$ 0.04 & 0.78 $\pm$ 0.02 & \textbf{0.83} $\pm$ \textbf{0.04} & 0.66 $\pm$ 0.04 \\
\midrule
$\psi$     & \textbf{0.74} $\pm$ \textbf{0.04} & \textbf{0.76} $\pm$ \textbf{0.03} & 0.78 $\pm$ 0.05  & 0.64 $\pm$ 0.04 & \textbf{0.79} $\pm$ \textbf{0.05} & - & \textbf{0.83} $\pm$ \textbf{0.04} &  0.66 $\pm$ 0.05 \\

$\psi$+Dropout    & \textbf{0.75} $\pm$ \textbf{0.03} & 0.68 $\pm$ 0.08 & \textbf{0.80} $\pm$ \textbf{0.04} & \textbf{0.65} $\pm$ \textbf{0.02} & \textbf{0.81} $\pm$ \textbf{0.03} & - & \textbf{0.83} $\pm$ \textbf{0.03} & 0.66 $\pm$ 0.04 \\
\bottomrule
\end{tabularx}
}}
\end{center}
\caption{In source models using psychophysical labels $\psi$ in an additional \textbf{regularization} term, we see improved test accuracy on classification tasks averaged across seeds on the \textcolor{imagenet-color}{Psych-ImageNet} and \textcolor{omniglot-color}{Psych-Omniglot} datasets (\textcolor{iam-color}{Psych-IAM} is pictured in Table~\ref{tab:two}). Each row denotes the regularization method utilized with multi-class cross-entropy loss. We computed error bars using standard error across 5 seeds. Columns with multiple bolded results indicate overlap between standard errors. In the psychophysical data point for VGG-16, the model always overfit the data here, yielding negligible results. VGG-16 has 138M parameters, compared to the 25M parameters for ResNet-50.}
\label{tab:one}
\end{table*}

Table~\ref{tab:one} demonstrates the main results of combining psychophysical labels with standard regularizers on ResNet-50, VGG-16, ViT, and PredNet on \textcolor{imagenet-color}{Psych-ImageNet} and \textcolor{omniglot-color}{Psych-Omniglot} datasets. We see general trends that models fine-tuned psychophysical labels perform better than others on the same task.
% more elaboration
Below the midline, we see a row $\psi$ and a row $\psi+$Dropout. The CNN models all improve more with either one of these (in some cases, the difference was minute, and both rows were bolded). Likewise, the high behavioral-fidelity model ViT also improved strongly, but the high activation-fidelity model PredNet did not show significant improvement in all but one case.

Likewise, we observe improved CER and WER on \textcolor{iam-color}{Psych-IAM} with psychophysical regularizers when compared to control groups in Table~\ref{tab:two}. Interestingly, only the neural model improves significantly when adding a psychophysical regularization term, perhaps because the annotator labels were biased more towards a neural model.

We use these findings in transfer learning evaluation~\ref{sub:transfer_learning_eval} to see how fine-tuning a model with psychophysical labels enables it to generalize better across different datasets.

\begin{table}
  \begin{center}
    % {\small{
    \textcolor{iam-color}{Psych-IAM} \\
\begin{tabular}{l|ccc}
Method                              & CER                  & WER  \\
\midrule
ResNet-50$+$CRNN                          & 0.15 $\pm$ 0.01  & 0.33 $\pm$ 0.02 \\
ViT                   & 0.13 $\pm$ 0.02  & 0.44 $\pm$ 0.01 \\
\midrule
ResNet-50$+$CRNN$+\psi$           & \textbf{0.11 $\pm$ 0.01}  & \textbf{0.31 $\pm$ 0.02} \\
ViT$+\psi$    & 0.14 $\pm$ 0.01  & 0.42 $\pm$ 0.02 \\

\bottomrule
\end{tabular}
% }}
\end{center}
\caption{We perform experimental runs on the \textcolor{iam-color}{Psych-IAM} dataset using a ResNet-50$+$CRNN and ViT architectures in the first two rows. In the bottom two rows, we regularize the loss function with psychophysical labels, represented by $\psi$. This improves both CER and WER on this dataset, suggesting further generalizability of psychophysical regularizers in fine-grained tasks. \emph{Lower is better.}}
\label{tab:two}
\end{table}

\textbf{Models with High Behavioral-Fidelity Improve More with Psychophysical Regularization.} 
Often, we see the trend that the more behavioral-fidelity models have stronger proportionate changes from the additional regularization. This suggests that the salient latent information encoded by the psychophysically-annotated datasets into neural models holds better generalization capacity than datasets that do not contain these labels. % maybe something about entropy and change between datasets

The results suggest that models with better behavioral fidelity, with improved inductive biases more \emph{aligned} with humans, improve more from loss functions that incorporate human behavioral measurements into the regularization term. 

In contrast, we see less of a gain from a psychophysical regularizer when using a high activation-fidelity model like PredNet. We further postulate that the model attunes more toward activation fidelity than toward outcome fidelity. Psychophysical labels do not possess activations --- they are a distribution of human behavioral responses to similar stimuli of what the model sees, but still higher level than neuronal activity. 

While these results are preliminary in that only a few models were tested in the zoo of models that exist, it shows that there may exist potential for using psychophysical regularizers to improve performance in generalizability in a variety of domains. 

\subsection{Transfer Learning Evaluation}
\label{sub:transfer_learning_eval}

In these experiments, we compared how a fine-tuned model on a psychophysically annotated dataset performed on a different dataset. We fine-tuned a ResNet-50, VGG-16, ViT, and PredNet~\footnote{PredNet was pre-trained on KITTI cities first. For more details on this, please see supp. mat.} on each psychophysically annotated dataset. From there, each fine-tuned model was run on a different dataset from what it was fine-tuned with. For each trial, each model was run 5 different times on each of the transfer learning tasks, with error bars taking the standard error among each the runs. 

\begin{figure}[t]
\centering
\includegraphics[width=1\linewidth]{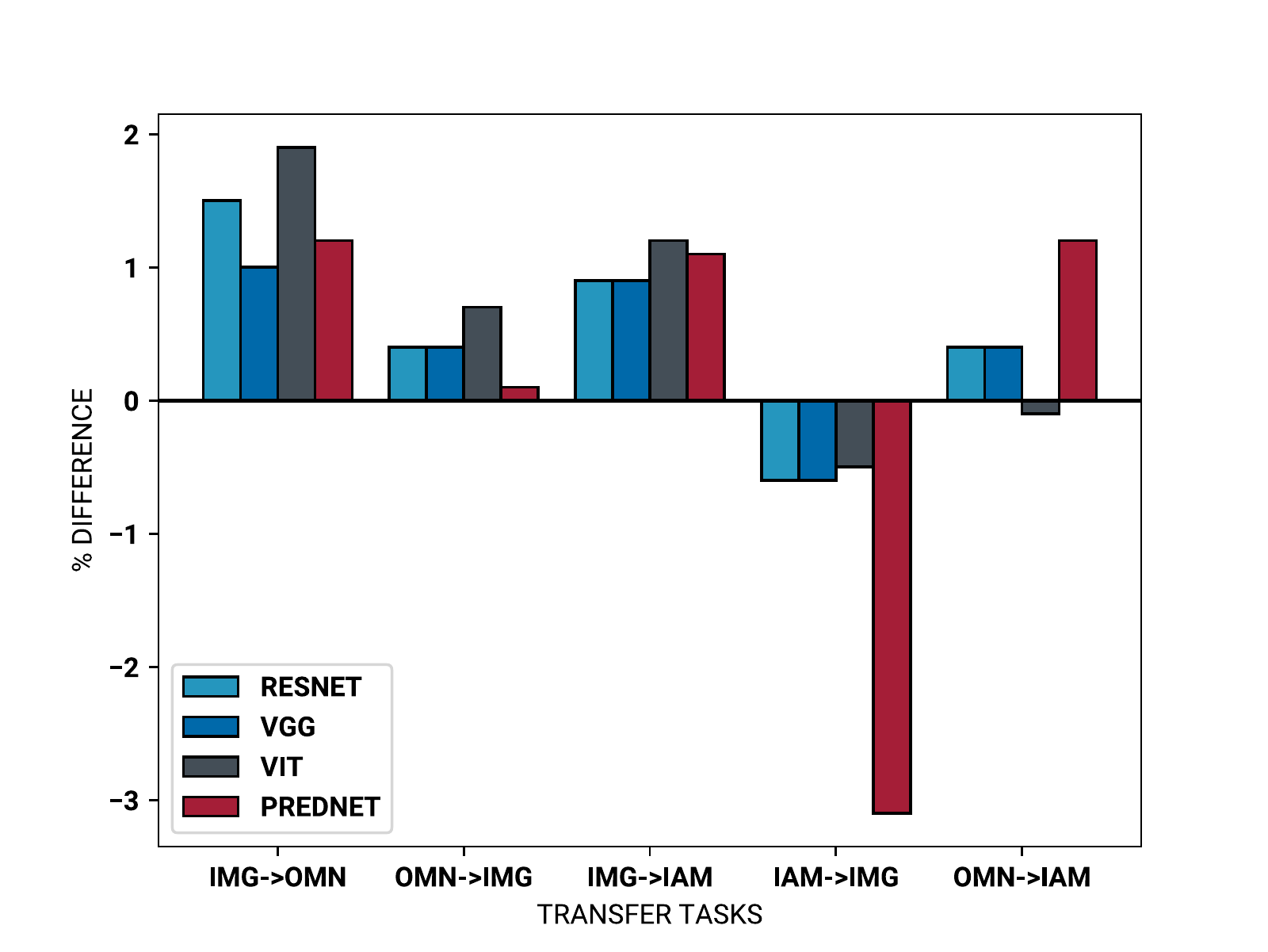}
\caption{The \% difference of average model performance across several transfer learning tasks when adding \modelName. Notice model performance generally improves across transfer learning tasks, but the models on the \textcolor{iam-color}{Psych-IAM}$\rightarrow$\textcolor{imagenet-color}{Psych-ImageNet} fair much worse. The information gained from images of documents does not contain enough discriminability to transfer to a more complex dataset. Just as Skinner postulated, transfer learning is still limited by tasks with strong perceptual differences~\cite{skinner2014contingencies}. In this case, it remains impossible to generalize from fine-grained handwriting data to generic object recognition.}
\label{fig:five}
\end{figure}

Figure~\ref{fig:five} demonstrates the potential of transfer learning of a psychophysically fine-tuned model towards another dataset. Overall, models that are trained on a more variant dataset provide greater transfer learning performance gains. 

In particular, models that were fine-tuned on \textcolor{imagenet-color}{Psych-ImageNet}, a high-variance dataset, transferred performance well. This is likely due to the fact that \textcolor{imagenet-color}{Psych-ImageNet} consists of hierarchical collections of object images; human participants, when providing annotations on this data, naturally recognize complex representations of objects more readily than most artificial systems~\cite{geirhos2018imagenet, dicarlo2012does} (see Supp. Mat. for more details on ablations).

% Justification in text of why it does worse in third group
In contrast, most models that were fine-tuned on datasets with lower complexity than the transfer objective did not see the same improvements. For example, the ResNet-50s fine-tuned on the \textcolor{iam-color}{Psych-IAM} handwriting tasks performed slightly worse on the \textcolor{imagenet-color}{Psych-ImageNet} dataset than models which did not receive this transfer tasks (grouping 3 in Figure~\ref{fig:five}). \modelName only remain helpful when human annotators, who handle cognitively valuable tasks, transfer to tasks of similar or less complex. Likewise, the distributions, in pixels and otherwise, between \textcolor{imagenet-color}{Psych-ImageNet} and \textcolor{iam-color}{Psych-IAM} are different. Furthermore, when flipping the transfer learning tasks, with models pre-trained on \textcolor{imagenet-color}{Psych-ImageNet} against the \textcolor{iam-color}{Psych-IAM} dataset, see a slight $\%$ increase instead of a loss. 

Qualitatively, a caveat to our method was finding the right tasks to transfer models among. We decided to keep the results transferring \textcolor{iam-color}{Psych-IAM}$\rightarrow$\textcolor{imagenet-color}{Psych-ImageNet} to offer a counter to the claim that \modelName works for every transfer learning task. In addition, each transfer task would improve with further hyperparameter optimization, but we held those constant for these experimental trials for consistency. 

\subsection{Brain-Score Evaluation}
\label{sub:brain-score}

Brain-Score is an effective metric for computing correlations among activations in biologically-inspired neural models by comparing them to neural activations in biological units, but it does not account for \emph{behavioral} differences. Higher brain-scores correlate with biological fidelity with respect to neural activations in the visual cortex~\cite{schrimpf2020brain}. 

There is also recent work~\cite{richardwebster2022perceptual} in measuring behavioral fidelity in neural models, but it fails to have a comprehensive suite of benchmarks like Brain-Score. In further analysis, we hope to use additional metrics beyond Brain-Score once they reach consistency. Additionally, we care about explicit activation fidelity with Brain-Score, so we do not utilize perceptual similarity metrics~\cite{zhang2018unreasonable, kumar2022better}.
The biological mechanism may differ from the neural activity aligned with the learning task, resulting in a lower Brain-Score than expected. 

% maybe squeeze this table to the right of the transfer learning one
% brain-score
\begin{table}
  \begin{center}
    {\small{
\begin{tabular}{lr}
Model                       & Brain-score \\
\toprule    
ResNet-50                   & 0.432 \\
ViT                         & 0.374 \\
PredNet                     & 0.182 \\
\midrule % Psych stuff below
$\psi+$ResNet-50            & \textbf{0.445} \\
$\psi+$ViT                 & 0.377 \\ 
$\psi+$PredNet             & 0.182 \\
\bottomrule
\end{tabular}
}}
\end{center}
\caption{\textbf{Brain-Score} for best models trained on \textcolor{imagenet-color}{Psych-ImageNet}. Brain-Score serves as a metric for neurological fidelity by measuring the activations of models and comparing them to neural activations in biological systems. \emph{Higher is better.}}
\label{tab:four}
\end{table}

This contradicts the observation that some models in our experiments saw greater gains from \modelName despite having a lower Brain-Score. 
In particular, PredNet has good transfer learning improvements when using additional psychophysical annotations, yet has an extremely low Brain-Score. We postulate that running PredNet on only one activation node, as opposed to a greater number of them, limited the amount of biologically-plausible activations for Brain-Score to measure with its task suite. 
These results may indicate that activation fidelity does not always correlate with behavioral fidelity.

\section{Conclusion, Limitations, Broader Impacts}
\label{sec:conclusion}
% Summary
We introduced \modelName as a method to generalize transfer learning tasks. We compared the performance of a variety of machine learning models with psychophysical regularizers among $\ell_{1}$-regularization, $\ell_{2}$-regularization, and Dropout as regularizers and a control group for each model with no regularizer present. Likewise, we tested model behaviors among different transfer learning tasks. Lastly, we used Brain-Score to compare neurological fidelity among learned models. \modelName performs better than traditional transfer learning, and we hope this work inspires further research into psychophysical research in the computer vision community. 
% We observed that \modelName best benefited PredNet as a complete framework. 

% Scale Limitation
Any use of \modelName is limited by the size and fidelity of its human-annotated dataset. In our experiments, we used data collected over simple Amazon Mechanical Turk trials. While we demonstrated that the annotations gathered in this space had transferred benefits to other models, the extent of this benefit could potentially change in different types of machine learning tasks. 
% \modelName utilizes human-annotated data, and this information yields no identifying personal information other than an individual's reaction time toward a forced-choice task. With the ease of use of Mechanical Turk and other companies that can easily collect and aggregate human-annotated data at scale, we anticipate further research and development into ways to use human annotations directly in the loss space. At a superficial level, transfer learning of psychophysical annotations for a classifier is akin to imitation learning. In feature studies, we hope to explore this paradigm.  
% Broader Impact
Despite limitations, crowd-sourced data is still simple to collect when compared to tightly controlled laboratory measurements while still improving model performance. In future works, we hope to see \modelName in other paradigms, such as reinforcement learning. 

Because \modelName remains easily scales without revealing much about the annotator, we do not anticipate adverse effects with it alone, but we caution for adversarial use cases that further prejudice and unjust biases. 
% We plan to release the code and dataset for this paper after its publication. 
% \section*{Acknowledgements}
% {\smaller\begin{spacing}{0.5}
% At vero eos et accusamus et iusto odio dignissimos ducimus qui blanditiis praesentium voluptatum deleniti atque corrupti quos dolores et quas molestias excepturi sint occaecati cupiditate non provident, similique sunt in culpa qui officia deserunt mollitia animi, id est laborum et dolorum fuga. Et harum quidem rerum facilis est et expedita distinctio. Nam libero tempore, cum soluta nobis est eligendi optio cumque nihil impedit quo minus id quod maxime placeat facere possimus, omnis voluptas assumenda est, omnis dolor repellendus. Temporibus autem quibusdam et aut officiis debitis\end{spacing}}

{\small
\bibliographystyle{ieee_fullname}
\bibliography{main.bib}
}

\appendix
\clearpage

%%% 

% %%%%%%%%% TITLE
% \title{Using Human Perception to Regularize Transfer Learning: Supplementary Material}

% \author{
% Justin Dulay and Walter J. Scheirer \\
% Dept. of Computer Science and Engineering\\
% University of Notre Dame\\
% {\tt\small \{jdulay, wscheire\}@nd.edu} \\
% % \and
% % Walter J. Scheirer$^\spadesuit$\\
% % Dept. of Computer Science and Engineering\\
% % $^\spadesuit$University of Notre Dame\\
% % {\tt\small walter.scheirer@nd.edu}
% }

% \maketitle

\begin{abstract}
    Here, we discuss additional details to the main \modelName paper. We organize it by:
    
    \begin{itemize}
        \item Dataset descriptions~\ref{sup:section:dataset}.
        \item PredNet implementation details~\ref{sup:section:prednet}.
        \item Ablations of other experiments~\ref{sup:section:model}.
    \end{itemize}
    
    The items in this supplementary material serve to provide additional detail and context to the main text while not adding or removing pertinent information from it. 
    
\end{abstract}

\section{Dataset Descriptions.}
\label{sup:section:dataset}

\subsection{\textcolor{imagenet-color}{Psych-ImageNet}}
\begin{itemize}
    \itemsep0em 
    \item The dataset has 293 known classes in total, excluding other open-set classes to use at later studies.
    \item There are 40 classes with psychophysical labels, producing a ratio of psychophysically-annotated to original classes as in~\cite{scheirer2014perceptual}.
    \item There are 33,548 known training samples in total, and 12,428 samples have corresponding reaction times. 
    \item Each data point has a reaction time, class label, and ImageNet-sized (224x224) image associated with it.
    \item Reaction times collected (each reaction time is the amount of time to choose a stimulus, given 5 other examples). Responses in this data were collected for class recognition against noisy stimuli. 
\end{itemize}

\begin{figure*}[ht]
\centering
\includegraphics[width=1\textwidth]{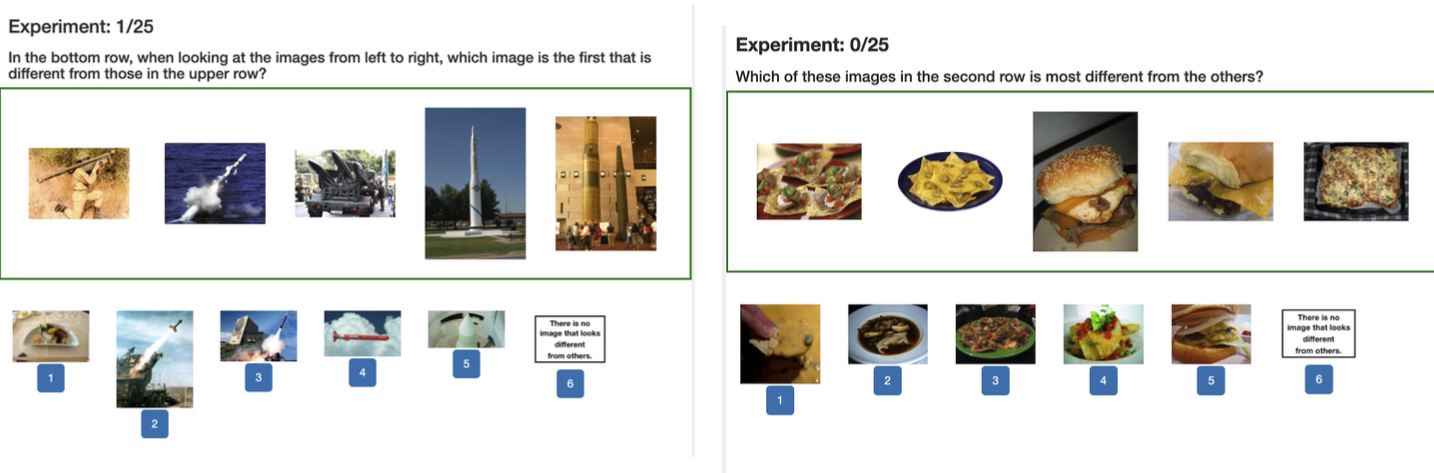}
\caption{\textbf{Crowd-sourced tasks from ~\cite{huang2022measuring}.} The above figure contains two screenshots from the worker data aggregator view in Amazon Mechanical Turk. The image on the left contains an \textcolor{red}{easy} example where most annotators answered quickly and accurately; a model that fails to answer in the same way receives a higher penalty. Likewise, the screenshot on the right contains a more \textcolor{blue}{difficult} class, where a model does not receive as harsh of a penalty for answering incorrectly.}
\label{sup:fig:one}
\end{figure*}

% task description
Each trial was an object-matching task, where the human participant of 5 opposed stimuli to it (see supplementary material). Each image was from one of the 293 Psych-ImageNet classes chosen for the task. The participant had to select the object they thought belonged to the top sample or rejected it together, should there be no match. A timer collected the participants reaction time for each question. Best viewed in color. An example of crowd-sourced task pairings from~\cite{huang2022measuring} is shown in Figure~\ref{sup:fig:one}.

Classes were evenly distributed across trials, as were positive vs negative matches. Likewise, the difficulty of experiments was variable to avoid a ceiling effect, a form of scale attenuation in which the maximum performance measured does not reflect the true maximum of the independent variable. More dataset details can be found in the supplementary material.

\subsection{\textcolor{omniglot-color}{Psych-Omniglot}}
\begin{figure}[t]
\centering
\includegraphics[width=1\linewidth]{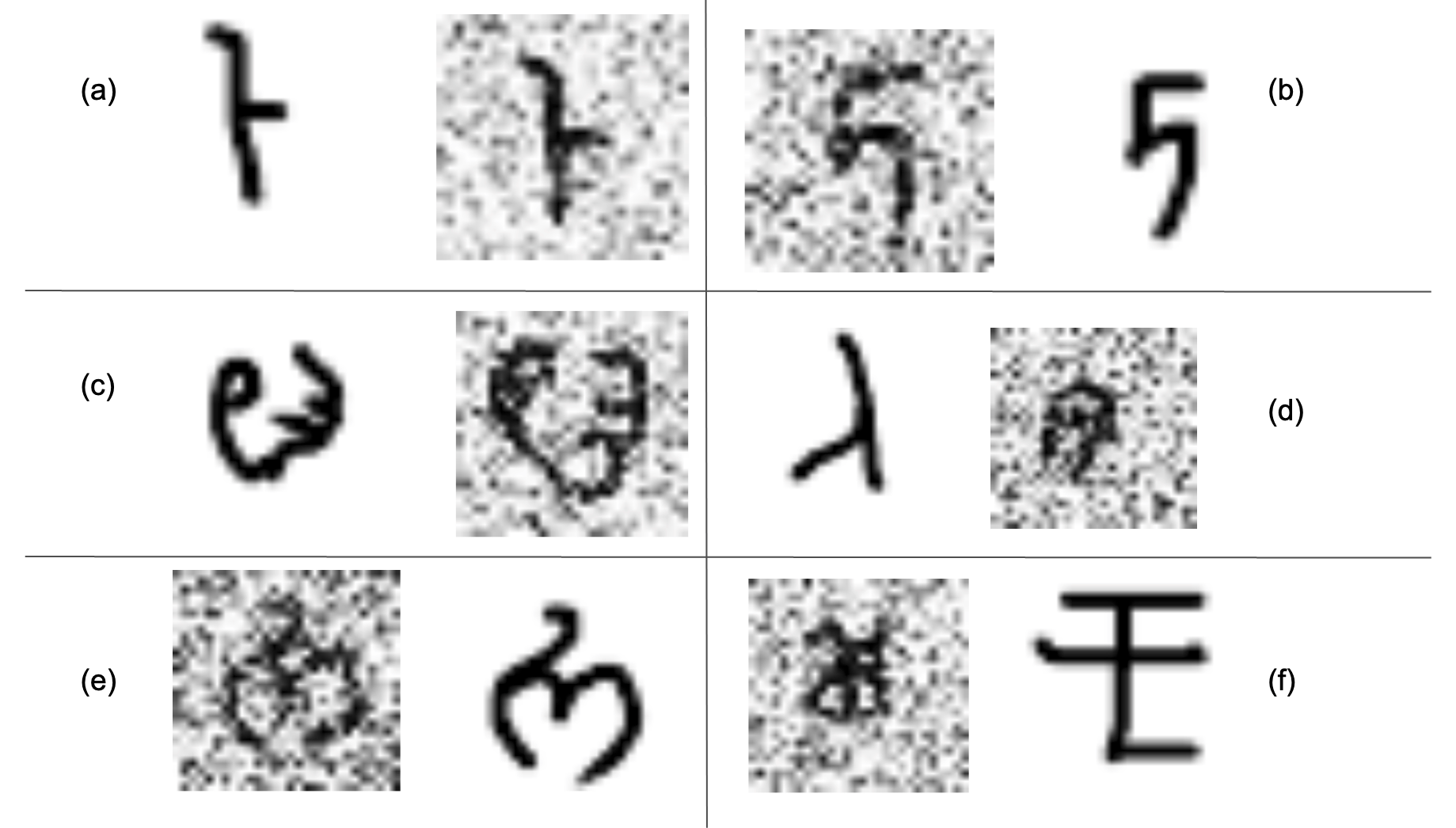}
\caption{\textbf{Crowd-sourced tasks from ~\cite{dulay2022guiding}.} An example two-alternative forced choice OCR task as seen from the participant's view. Labels (d) and (f) represent character pairs where the class labels differ; the rest represent the same class pairing. 
The blurred and noisy images lead to more informative psychophysical labels for operationalization within the machine learning task during training.}
\label{sup:fig:two}
\end{figure}

The \textcolor{omniglot-color}{Psych-Omniglot} is a variant on the Omniglot dataset~\cite{lake2015human} with psychophysical labels collected from the research in ~\cite{dulay2022guiding}. The dataset contains images of handwritten characters from hundreds of typesets, many of which a typical crowd-sourced study participant would be unfamiliar with. The data is augmented with counterpart samples for each image with a deep convolutional generative adversarial network (DCGAN)~\cite{goodfellow2014generative} to increase intraclass variance and the sample size per class --- all of which are forms of implicit regularization. An example of crowd-sourced task pairings from~\cite{dulay2022guiding} is shown in Figure~\ref{sup:fig:two}.

In this dataset, human behavioral measurements were gathered as reaction times to stimuli in crowd-sourced experiments. Human participants were presented with two opposing stimuli from the original Omniglot dataset (a Two-Alternative Forced Choice task) and decided whether the two stimuli were the same character in the dataset. The reaction time from the participants was recorded automatically. Broadly speaking about the dataset as a whole, these human reaction times were long on hard pairings, and short on easy character pairings. The introduction of this easy vs. hard pairing would prove useful for supervised learning tasks. 

\subsection{\textcolor{iam-color}{Psych-IAM}}

The dataset is a modification of the IAM dataset~\cite{marti2002iam} with human behavioral measurements on lines of text collected from ~\cite{grieggs2021measuring} on about $35\%$ of the dataset (2,152 lines). 

In the main text, we report both word error rate (WER) and character error rate (CER) for this dataset. The word error rate is a model's error with respect to the individual word on the line in the dataset, while the character error rate corresponds to the model's fidelity with the human annotator's marks on the individual word. 

The reaction time of the annotator to accurately record a character and line was recorded in this annotated dataset~\cite{grieggs2021measuring}. For the main text study, we used the reaction times in conjunction with images of the text itself to perform transfer learning OCR tasks with \modelName. 

% dataset size justification
\subsection{Dataset Limitations}
We recognize that \textcolor{imagenet-color}{Psych-ImageNet} only contains annotations on 40 of the total 293 classes. While previous psychophysics and machine learning studies suggest that this still remains representative of the entire training distribution~\cite{scheirer2014perceptual, grieggs2021measuring}, reaction times on more classes may potentially yield better results. 

Likewise, \textcolor{omniglot-color}{Psych-Omniglot} is a dataset in which the annotations were collected \emph{via} Amazon Mechanical Turk. While the practitioners accounted for systematic errors, no crowd-sourcing study is entirely robust to untrustworthy annotators~\cite{stewart2017crowdsourcing}.

Lastly, \textcolor{iam-color}{Psych-IAM}, along with the other two datasets, also suffer from limits of overall numbers of available annotations due to academic budget constraints.

In spite of limitations, we show~\ref{sec:results} that even with a small level of annotations, model performance still generally improves. 

\section{PredNet Fine-Tuning.}
\label{sup:section:prednet}

\textbf{Formulation.}
psychophysical transfer learning pulls similar results on the evaluative steps, as well.

% prednet figure
\begin{figure}[t]
\centering
\includegraphics[width=.75\linewidth]{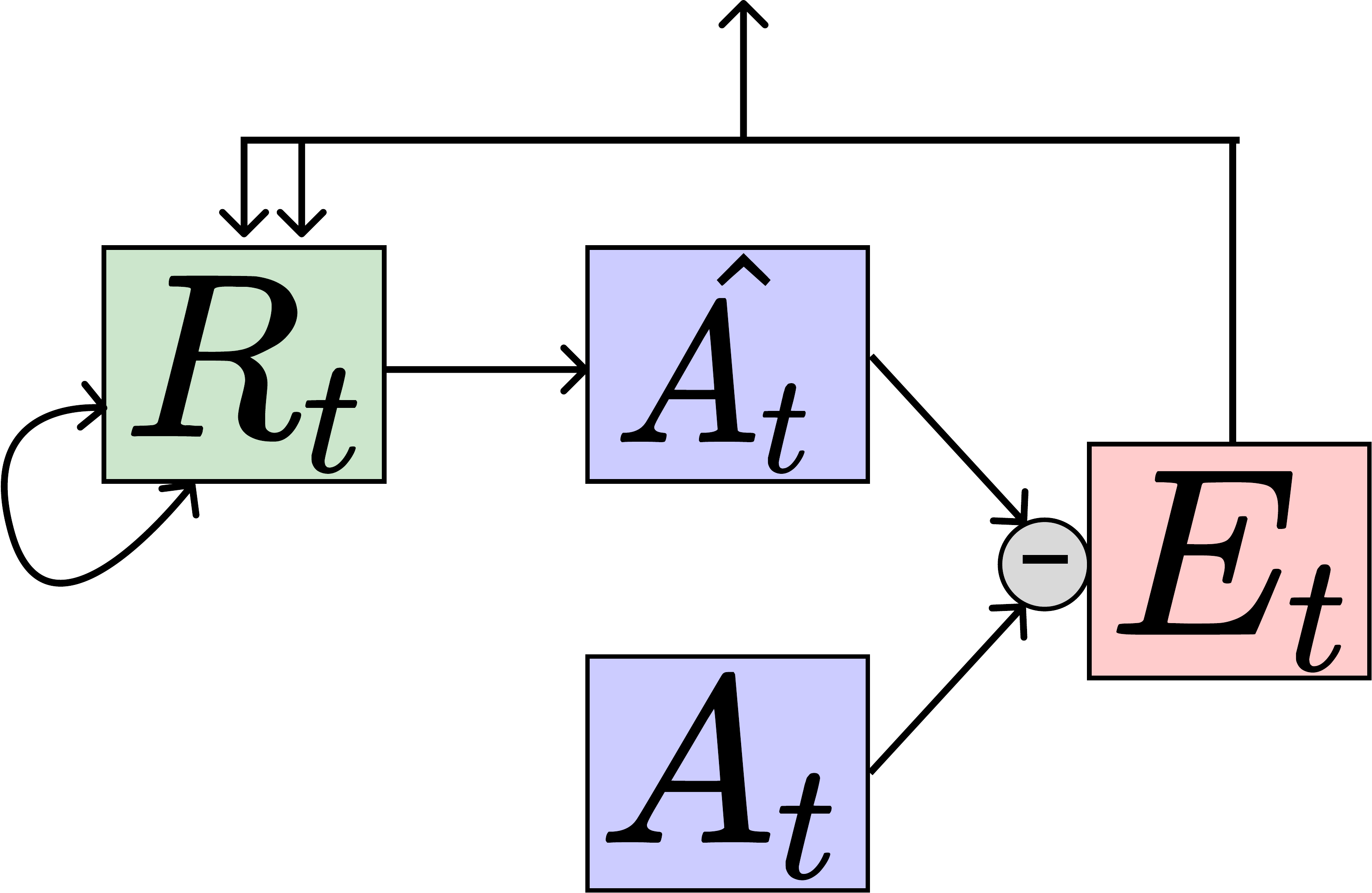}
\caption{The PredNet\cite{lotter2016deep} architecture.}
% The goal of this was to create a task space that challenges the subject --- if the task is too easy, they tend resort to overconfident guessing; if the task is too hard, they tend to give up on the task. The Gaussian noise perturbation was the most effective at increasing classification accuracy when psychophysically informing its feature representation space.}
\label{fig:two}
\end{figure}

Similarly to Blanchard et al.\cite{blanchard2019neurobiological}, we extract the activations of PredNet after the convLSTM layers. PredNet works with temporal data. First, We pre-trained it on videos from the KITTI~\cite{geiger2013vision} self-driving dataset. We set up the annotated Psych-ImageNet in an order of fixed frames and record the activations of PredNet at the fixed time steps. This representation at each neuron works like a supervised model's neuron in that we can add psychophysical transfer learning to it to regularize the learning representation.

The loss defined by PredNet is as follows: 

\begin{equation}
    \mathcal{L}_{t r a i n}=\sum_{t} \lambda_{t} \sum_{l} \frac{\lambda_{l}}{n_{l}} \sum_{n_{l}} E_{l}^{t}
    \label{equation:four}
\end{equation}

where $\lambda_{t}$ is a regularizer at the time step, $\frac{\lambda_{l}}{n_{l}}$ is a regularizing factor at a given layer in the network, and $E_{l}^{t}$ is the error at a time step~\ref{fig:two}. 

Indeed, the loss formulation for PredNet is inherently more complex than cross-entropy loss variations. For brevity, we conducted experiments to understand in which term should the psychophysical regularization variables be used. Again, each of the three terms uses a form of $\ell_{1}$-normalization to adjust model learning generalization. 

% table on term placement of regularizing loss
% PredNet parameter table can just go in the supp mat. add the actual results table back in to compare with the others, including maybe some transfer learning results with PredNet

% but lambda term ideas seem confusing when looking at all of this 
We observe that multiplying psychophysical transfer learning into the layer term $\lambda_t$ --- with the variable $\hat{A}_t$ yields the best results. In other words, after successive outputs of each layered convLSTM, we see performance gains more vividly than any other term within this loss. Furthermore, the regularization effect of psychophysical transfer learning, the softening of sharp gradient turns, pronounces the most at longer time steps on average (\emph{e.g. at steps $>5$}). As table ~\ref{sup:tab:one} suggests, the loss mostly benefited from psychophysical transfer learning regularization on the outputs between step outputs $\hat{A}_t$ at times $t$ ~\ref{sup:tab:two}. 

% moved that figure to the supp mat - table is informative enough
This result suggests that predictive coding networks in some way manage the latent ideals encoded in the psychophysical transfer learning data. 

While this experiment step was not part of the \emph{model-evaluative}, we believed it important to fine-tune psychophysical transfer learning on a non-traditional loss framework before conducting experimentation on the relative effects of psychophysical transfer learning on model-evaluative performance.

The pre-training of PredNet and the subsequent transfer to task to a frame-by-frame prediction on the modified Psych-ImageNet allows for the beneficial usage of psychophysical transfer learning. While this case is a niche, it demonstrates the viability of utilizing psychophysical transfer learning in a variety of future neurologically-inspired models.

\begin{table}
  \begin{center}
    {\small{
\begin{tabular}{lr}
\textcolor{imagenet-color}{Psych-ImageNet} & MAE \\
Method                   & PredNet \\
\toprule
Control                  & 0.59 $\pm$ 0.03 \\
$\ell_{1}$               & 0.62 $\pm$ 0.02 \\
$\ell_{2}$               & 0.61 $\pm$ 0.03 \\
Dropout                  & 0.61 $\pm$ 0.05 \\
Dropout+$\ell_{1}$       
& 0.61 $\pm$ 0.02 \\
\midrule
$RegularPsych$             & 0.64 $\pm$ 0.04 \\
$RegularPsych$+Dropout     & \textbf{0.65} $\pm$ \textbf{0.02} \\
\bottomrule
\end{tabular}
}}
\end{center}
\caption{On models using $RegularPsych$ as an evaluator, we see improved mean squared error reduction. All models were pre-trained on KITTI and evaluated on the house dataset. We computed error bars using standard error across 5 seeds. \emph{Lower is better.}}
\label{sup:tab:one}
\end{table}

\begin{table}
  \begin{center}
    {\small{
    
\begin{tabular}{lr|r}
\textcolor{imagenet-color}{Psych-ImageNet} && \textcolor{omniglot-color}{Psych-Omniglot} \\

\toprule
Parameter & Train Error & Train Error \\
\midrule
None & 0.12 $\pm$ 0.03 & 0.20 $\pm$ 0.05 \\
$A_t$ & 0.11 $\pm$ 0.04 & 0.19 $\pm$ 0.05\\
$\hat{A_t}$ & 0.06 $\pm$ 0.02 & 0.17 $\pm$ 0.04 \\

\bottomrule
\end{tabular}
}}
\end{center}
\caption{The table shows the train errors for each parameter selection of \emph{which} PredNet architecture layer to multiply by the $RegularPsych$ variable. The None column assumes a cross-entropy loss without any modification to the PredNet loss. The input layer $A_t$ shows no significant change in performance, regardless of what the psychophysical annotations are. However, we see a significant reduction in training error when applying $RegularPsych$ to the model prediction logits $\hat{A_t}$.}
\label{sup:tab:two}
\end{table}

% \section{Regularization specifics}

% \textcolor{red}{l1}

% \textcolor{red}{l2}

% \textcolor{red}{Dropout}

% \section{stuff on transfer learning}
\newpage
\section{Model Ablations}
\label{sup:section:model}

In Table~\ref{sup:tab:three}, we report ablation results on the \modelName transfer learning tasks. These show some additional transfer learning movements among different tasks in the experiments. For example, the first row of the figure represents the task shift, where the color of the $\psi$ represents the domain the psychophysical labels were gathered on. Not all domains transfer well, but there exist several domains where transfer learning works naturally. 

In this work, it remains apparent that the object recognition task and psychophysical labels from models learned on ~\textcolor{imagenet-color}{Psych-ImageNet} transfer well to the other domains used in this study. In rows $1$ and $3$ in Table~\ref{sup:tab:three}, we see the largest gains supported by this. Likewise, the transfer of domains from ~\textcolor{iam-color}{Psych-IAM} character annotation tasks to generic object recognition, in line with commonsense, does not transfer well. 

In future studies, we plan to explore different learning paradigms (e.g. reinforcement learning) to expand the results of transfer among domains. 

\begin{table*}
    % {\small{
        \begin{tabularx}{\textwidth}{l|X|X|X|X}
            & orig. + new + \%diff & orig. + new + \%diff & orig. + new + \%diff & orig. + new + \%diff \\
            Transfer Task                                                                       & ResNet  & VGG  & ViT & PredNet \\
            \midrule
            
            % ImageNet                     $\rightarrow$ \textcolor{imagenet-color}{Psych-ImageNet} & acc1ddd+acc2ddd+1.2\%.& acc1ddd+acc2ddd+1.2\% &  acc1ddd+acc2ddd+1.2\% \\
            % Omniglot                     $\rightarrow$ \textcolor{omniglot-color}{Psych-Omniglot}   & 1.0\%     & - &  1.9\%\\
            % \textcolor{red}{IAM}                          $\rightarrow$ \textcolor{iam-color}{Psych-IAM}        & -0.2\%    & - &  1.0\% \\
            % \midrule
            \textcolor{imagenet-color}{$\psi$} $\rightarrow$ \textcolor{omniglot-color}{$\psi$} &
            {\begin{tabularx}{\linewidth}{lll} 0.79&0.81&+1.5\% \end{tabularx}} &
            - & 
            {\begin{tabularx}{\linewidth}{lll} 0.83&0.85&\textbf{+1.9\%} \end{tabularx}} &
            {\begin{tabularx}{\linewidth}{lll} 0.63&0.65&+1.2\% \end{tabularx}}  \\
            \textcolor{omniglot-color}{$\psi$} $\rightarrow$ \textcolor{imagenet-color}{$\psi$} & 
            {\begin{tabularx}{\linewidth}{lll} 0.74&0.75&+0.4\% \end{tabularx}} & 
            {\begin{tabularx}{\linewidth}{lll} 0.76&0.76&+0.4\% \end{tabularx}} & 
            {\begin{tabularx}{\linewidth}{lll} 0.78&0.79&+0.7\% \end{tabularx}} & {\begin{tabularx}{\linewidth}{lll} 0.65&0.65&+0.1\% \end{tabularx}} \\
            \textcolor{imagenet-color}{$\psi$} $\rightarrow$ \textcolor{iam-color}{$\psi$} & 
            {\begin{tabularx}{\linewidth}{lll} 0.91&0.92&+0.9\% \end{tabularx}} & 
            {\begin{tabularx}{\linewidth}{lll} 0.81&0.02&-0.5\% \end{tabularx}} & 
            {\begin{tabularx}{\linewidth}{lll} 0.86&0.88&+1.2\% \end{tabularx}} & 
            {\begin{tabularx}{\linewidth}{lll} 0.64&0.65&+1.1\% \end{tabularx}}  \\
            \textcolor{iam-color}{$\psi$} $\rightarrow$ \textcolor{imagenet-color}{$\psi$} & 
            {\begin{tabularx}{\linewidth}{lll} 0.74&0.73&-0.6\% \end{tabularx}} & 
            {\begin{tabularx}{\linewidth}{lll} 0.76&0.76&+0.2\% \end{tabularx}} & 
            {\begin{tabularx}{\linewidth}{lll} 0.78&0.77&-0.5\% \end{tabularx}} & {\begin{tabularx}{\linewidth}{lll} 0.65&0.62&-3.1\% \end{tabularx}} \\
            \textcolor{omniglot-color}{$\psi$} $\rightarrow$ \textcolor{iam-color}{$\psi$} & 
            {\begin{tabularx}{\linewidth}{lll} 0.91&0.91&+0.4\% \end{tabularx}} & 
            {\begin{tabularx}{\linewidth}{lll} 0.81&0.81&+0.1\% \end{tabularx}} & 
            {\begin{tabularx}{\linewidth}{lll} 0.86&0.86&-0.1\% \end{tabularx}} & 
            {\begin{tabularx}{\linewidth}{lll} 0.65&0.66&+1.2\% \end{tabularx}}   \\
            \bottomrule
        \end{tabularx}
    % }}
\\
\caption{\textbf{Transfer learning $\%$ difference table.} With psychophysical transfer learning, performance increases by as much as \textbf{1.9$\%$}. Each row represented in this table represents a difference transfer learning task, denoted by $\psi$ corresponding in color with the dataset used. Each trial is the standard error across 5 seeds.
\emph{Note: using accuracy as $1-CER$ on \textcolor{iam-color}{Psych-IAM} trials in this table. Higher is better.}}
\label{sup:tab:three}
\end{table*}

% \newpage

% {\small
% \bibliographystyle{ieee_fullname}
% \bibliography{supplement}
% }

\end{document}